\documentclass[runningheads]{llncs}
\usepackage{graphicx}

\usepackage{tikz}
\usepackage{comment}
\usepackage{amsmath,amssymb} 
\usepackage{color}

\usepackage{booktabs} 
\usepackage{multirow}
\usepackage{pifont}
\usepackage[belowskip=-10pt,aboveskip=0pt]{caption}
\begin{document}
\pagestyle{headings}
\mainmatter
\def\ECCVSubNumber{2129}  

\title{Action Localization through Continual Predictive Learning} 


\titlerunning{Action Localization through Continual Predictive Learning}
%
\author{Sathyanarayanan N. Aakur$^1$ \and
Sudeep Sarkar$^2$}
\authorrunning{SN. Aakur et al.}
%
\institute{Oklahoma State University, Stllwater, OK, 74074 \\
\email{saakurn@okstate.edu} \and
University of South Florida, Tampa, FL 33620 \\
\email{sarkar@usf.edu}\\
}
\maketitle

\begin{abstract}

The problem of action recognition involves locating the action in the video, both over time and spatially in the image. The dominant current approaches use supervised learning to solve this problem, and require large amounts of annotated training data, in the form of frame-level bounding box annotations around the region of interest. In this paper, we present a new approach based on continual learning that uses feature-level predictions for self-supervision. It does not require any training annotations in terms of frame-level bounding boxes. The approach is inspired by cognitive models of visual event perception that propose a prediction-based approach to event understanding. We use a stack of LSTMs coupled with CNN encoder, along with novel attention mechanisms, to model the events in the video and use this model to predict high-level features for the future frames. The prediction errors are used to continuously learn the parameters of the models.  This self-supervised framework is not complicated as other approaches but is very effective in learning robust visual representations for both labeling and localization. It should be noted that the approach outputs in a streaming fashion, requiring only a single pass through the video, making it amenable for real-time processing. We demonstrate this on three datasets - UCF Sports, JHMDB, and THUMOS'13 and show that the proposed approach outperforms weakly-supervised and unsupervised baselines and obtains competitive performance compared to fully supervised baselines. Finally, we show that the proposed framework can generalize to egocentric videos and obtain state-of-the-art results in unsupervised gaze prediction. 

\end{abstract}

\section{Introduction}
We develop a framework for jointly learning spatial and temporal localization through continual, self-supervised learning, in a streaming fashion, requiring only a single pass through the video. 
Visual understanding tasks in computer vision have focused on the problem of recognition~\cite{kuehne2014language,karpathy2014large,aakurCRV2017,aakur2019wacv} and captioning~\cite{aakur2019wacv,venugopalan2014translating,venugopalan2015sequence,guo2016attention}, with the underlying assumption that each input video is already localized both spatially and temporally. 
While there has been tremendous progress in action localization, it has primarily been driven by the dependence on large amounts of tedious, spatial-temporal annotations. 
In this work, we aim to tackle the problem of spatial-temporal segmentation of streaming videos in a continual, self-supervised manner, without any training annotations.

\begin{figure*}[h]
\centering
\includegraphics[width=0.99\textwidth]{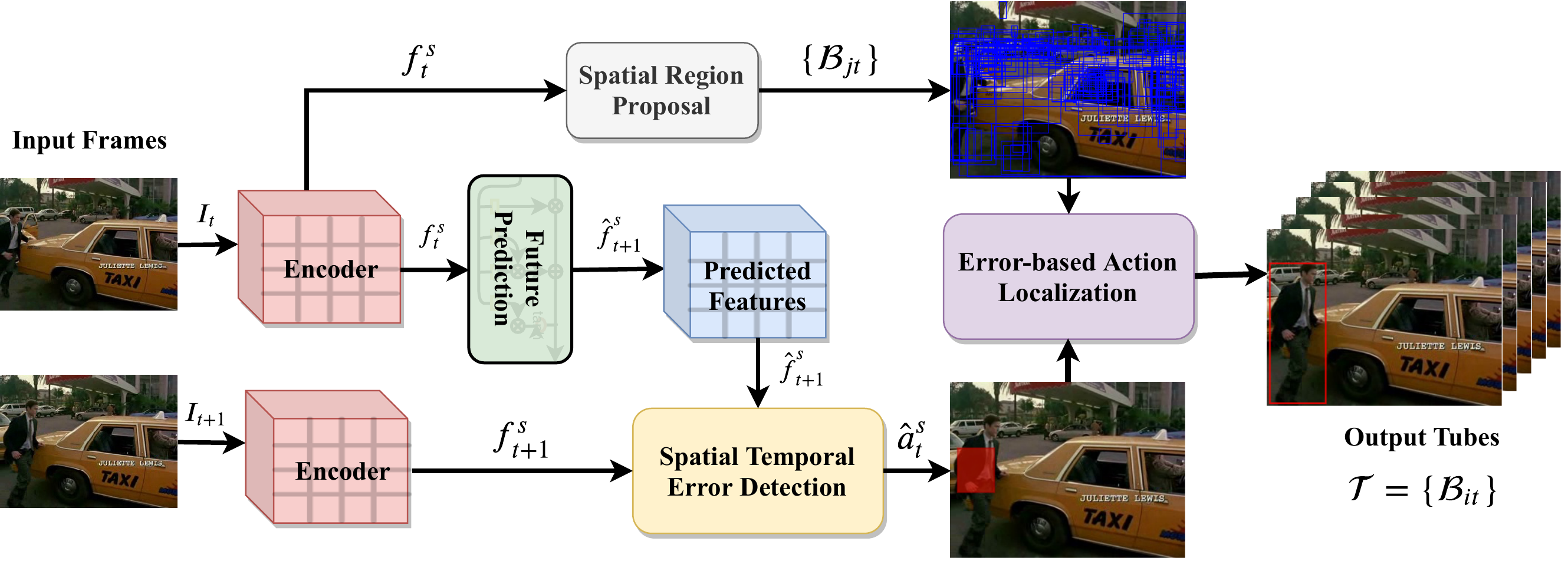}
\caption{The \textbf{Proposed Approach} has four core components: (i) feature extraction and spatial region proposal, (ii) a future prediction framework, (iii) a spatial-temporal error detection module and (iv) the error-based action localization process. 
} 
\label{fig:overallApproach}
\end{figure*}

Drawing inspiration from psychology \cite{horstmann2015surprise,horstmann2016novelty,zacks2001perceiving}, we consider the underlying mechanism for both event understanding and attention selection in humans as the idea of \textit{predictability}. 
Defined as the surprise-attention hypothesis~\cite{horstmann2015surprise}, unpredictable factors such as large changes in motion, appearance, or goals of the actor have a substantial effect on the event perception and human attention. 
Longer-term, temporal surprise has shown to have a strong correlation with event boundary detection~\cite{zacks2001perceiving,Aakur_2019_CVPR}, whereas short-term spatial surprise (such as those caused by motion) have a stronger effect on human attention and localization~\cite{horstmann2016novelty}.
Our approach combines both spatial and temporal surprise to formulate a computational framework to tackle the problem of self-supervised action localization in streaming videos in a continual manner. 

We formulate our computational framework on the idea of spatial-temporal feature anticipation for modeling predictability of perceptual features. The main assumption in our framework is that expected, unpredictable features require attention and often point to the actor performing the action of interest, whereas predictable features can belong to background clutter not relevant to the action of interest. It is to be noted that unpredictability or \textit{surprise} is not the same as \textit{rarity}. It refers to short-term changes that aid in the completion of an overall task, which can be recurring. We model the perceptual features using a hierarchical, cyclical, and recurrent framework, whose predictions are influenced by current and prior observations as well as current perceptual predictions. Hence, the predictive model’s output can influence the perception of the current frame being observed. The predictions are constantly compared with the incoming observations to provide self-supervision to guide future predictions. 

We leverage these characteristics to derive and quantify spatial-temporal predictability. Our framework performs continuous, prediction-based learning to generate ``\textit{attention maps}'' that are consistent with the predictability of the observed scene. With the attention derived from spatial-temporal unpredictability, we leverage advances in region proposals~\cite{redmon2016you,liu2016ssd,uijlings2013selective,zhu2015tracking} to localize actions in streaming videos without any supervision. Contrary to other attention-based approaches~\cite{escorcia2020guess,li2018videolstm,sharma2015action}, we do not use the object-level characteristics such as label, role, and affordance in the proposal generation process. 

\textbf{Contributions:} The contributions of our approach are three-fold: (i) we are among the first to tackle the problem of self-supervised action localization in \textit{streaming videos} without any training data such as labels or bounding boxes, (ii) we show that modeling spatial-temporal prediction error can yield consistent localization performance across action classes and (iii) we show that the approach generalizes to egocentric videos and achieves competitive performance on the \textit{unsupervised gaze prediction} task. 
\section{Related Work}
\textbf{Supervised action localization} approaches tackle the action localization problem through the simultaneous generation of bounding box proposals and labeling each bounding box with the predicted action class. Both the bounding box generation and labeling are fully supervised, i.e., they require ground truth annotations of both bounding boxes and labels. The typical approach is to leverage advances in object detection to include temporal information~\cite{gkioxari2015finding,tian2013spatiotemporal,wang2014video,jain2014action,tran2012max,soomro2016predicting,soomro2015action} for proposal generation. The final step typically involves the use of the Viterbi algorithm~\cite{gkioxari2015finding} to link the generated bounding boxes across time.

\textbf{Weakly-supervised action localization} approaches~\cite{li2018videolstm,escorcia2020guess,lan2011discriminative,sharma2015action} have been explored to reduce the need for extensive annotations. They typically only require video-level labels and rely on object detection-based approaches to generate bounding box proposals. It is to be noted that weakly supervised approaches also use object-level labels and characteristics to guide the bounding box selection process. Some approaches~\cite{escorcia2020guess} use a similarity-based tracker to connect bounding boxes across time to incorporate temporal consistency into the generation process.

\textbf{Unsupervised action localization} approaches have not been explored to the same extent as supervised and weakly-supervised approaches. These approaches do not require any supervision - both labels or bounding boxes. The two more common approaches are to generate action proposals using (i) super-voxels~\cite{jain2014action,soomro2017unsupervised} and (ii) clustering motion trajectories~\cite{van2015apt}. It should be noted that ~\cite{soomro2017unsupervised} also uses object characteristics to evaluate the ``humanness'' of each super-voxel to select bounding box proposals. 
Our approach falls into the class of unsupervised action localization approaches. The most closely related approaches (with respect to architecture and theme) to ours are VideoLSTM~\cite{li2018videolstm} and Actor Supervision~\cite{escorcia2020guess}, which use attention in the selection process for generating bounding box proposals, but require video-level labels. 
We, on the other hand, do not require any labels or bounding box annotations for training. 

While fully supervised approaches return more precise localization and achieve better recognition, the required number of annotations is rather large and is not amenable to an increase in the number of classes and a decrease in the number of training videos. Weakly supervised approaches, while not requiring frame-level annotations, have the underlying assumption that there exists a large, annotated training set that allows for effective detection of all possible actors (both human and non-human) in the set of action classes. Unsupervised approaches, such as ours, do not make any such assumptions but can result in poorer localization performance. We alleviate this to an extent by leveraging advances in region proposal mechanisms and learning robust representations for obtaining video-level labels. 

\section{Self-Supervised Action Localization}
In this section, we introduce our self-supervised action localization framework, as illustrated in Figure~\ref{fig:overallApproach}. Our approach has four core components: (i) feature extraction and spatial region proposal, (ii) a self-supervised future prediction framework, (iii) a spatial-temporal error detection module, and (iv) the error-based action localization process. 
\subsection{Feature Extraction and Spatial Region Proposal}\label{sec:proposal}
The first step in our approach is feature extraction and the subsequent \textit{per-frame} region proposal generation for identifying possible areas of actions and associated objects. Considering the tremendous advances in deep learning architectures for learning robust spatial representations, we use pre-trained convolutional neural networks to extract the spatial features for each frame in the video.
We use a region proposal module, based on these spatial features, to predict possible action-agnostic spatial locations. 
We use class-agnostic proposals (i.e., the object category is ignored, and only feature-based localizations are taken into account) at this stage for two primary reasons. First, we do not want to make any assumptions on the actor's characteristics, such as label, role, and affordance. Second, despite significant progress in object detection, there can be a lot of missed detections, especially when the object (or actor) performs actions that can transform their physical appearance. 
It is to be noted that these considerations can result in a large number of region proposals which require careful and robust selection but can yield higher chances of correct localization. 

\subsection{Self-supervised Future Prediction}
The second stage in our proposed framework is the self-supervised future prediction framework. 
We consider the future prediction module to be a generative model whose output is conditioned on two factors - the current observation and an \textit{internal event model}. The current observation $f^S_t$ is the feature-level encoding of the presently observed frame $I_t$. We use the same feature encoder as the region proposal module to reduce the memory footprint and complexity of the approach. 
The \textit{internal event model} is a set of parameters that can effectively capture the spatial-temporal dynamics of the observed event. Formally, we define the predictor model as $P(\hat{f}^S_{t+1}|W_e, f^S_t)$, where $W_e$ represents the internal event model and $\hat{f}^S_{t+1}$ is the predicted features at time $t+1$. Note that features $f^S_t$ is not a one-dimensional vector, but a tensor (of dimension $w_f \times h_f \times d_f$) representing the features at each spatial location. 

We model temporal dynamics of the observed event using Long Short Term Memory Networks (LSTMs)\cite{hochreiter1997long}. While other approaches~\cite{jia2016dynamic,vondrick2016anticipating,vondrick2017generating} can be used for prediction, we consider LSTMs to be more suited for the following reasons. First, we want to model the temporal dynamics across \textit{all} frames of the observed action (or event). Second, LSTMs can allow for multiple possible futures and hence will not tend to average the outcomes of these possible futures, as can be the case with other prediction models. Third, since we work with error-based localization, using LSTMs can ensure that the learning process propagates the spatial-temporal error across time and can yield progressively better predictions, especially for actions of longer duration.
Formally, we can express LSTMs as 
\begingroup
\allowdisplaybreaks
\begin{align}
\label{eqn:LSTM}
i_t = \sigma(W_{i}x_t + W_{hi}h_{t-1} + b_i) \\ 
f_t = \sigma(W_{f}x_t + W_{hf}h_{t-1} + b_f ) \\ 
o_t = \sigma(W_{o}x_t + W_{ho}h_{t-1} + b_o ) \\ 
g_t = \phi(W_{g}x_t + W_{hg}h_{t-1} + b_g ) \\
m_t = f_t \cdot m_{t-1} + i_t \cdot g_t \\
h_t = o_t \cdot \phi(m_t) 
\end{align}
\endgroup
where $x_t$ is the input at time $t$, $\sigma$ is a non-linear activation function, ($\cdot$) represents element-wise multiplication, $\phi$ is the hyperbolic tangent function (\emph{tanh}) and $W_{k}$ and $b_{k}$ represent the trained weights and biases for each of the gates.

As opposed to~\cite{Aakur_2019_CVPR}, who also use an LSTM-based predictor and a decoder network, we use a hierarchical LSTM model (with three LSTM layers) as our event model. This modification allows us to model both spatial and temporal dependencies, since each higher-level LSTMs act as a progressive decoder framework that captures the temporal dependencies captured by the lower-level LSTMs. The first LSTM captures the spatial dependency that is propagated up the prediction stack. 
The updated hidden state of the first (bottom) LSTM layer ($h^1_t$) depends on the current observation $f^S_t$, the previous hidden state ($h^1_{t-1}$) and memory state ($m^1_{t-1}$).
Each of the higher-level LSTMs at level $l$ take the output of the bottom LSTM's output $h^{l-1}_{t}$ and memory state $m^{l-1}_{t}$ and can be defined as $(h^{l}_{t}, m^{l}_{t}) = LSTM(h^{l}_{t-1}, h^{l-1}_{t}, m^{l-1}_{t})$. Note this is different from a typical hierarchical LSTM model~\cite{song2017hierarchical} in that the higher LSTMs are impacted by the output of the lower level LSTMs at current time step, as opposed to that from the previous time step. Collectively, the event model $W_e$ is described by the learnable parameters and their respective biases from the hierarchical LSTM stack.

Hence, the top layer of the prediction stack acts as the decoder whose goal is to predict the next feature $f^S_{t+1}$ given all previous predictions $\hat{f}^S_{1}, \hat{f}^S_{2}, \ldots \hat{f}^S_{t}$, an event model $W_e$ and the current observation $f^S_{t}$. We model this prediction function as a log-linear model characterized by

\begin{equation}
\log\,p(\hat{f}^{s}_{t+1} | h^l_t) = 
\sum_{n=1}^{t} f(W_e, f^S_t) + log\,Z(h_t)
\label{eqn:predictedFeat}
\end{equation}
where $h^l_t$ is the hidden state of the $l^{th}$ level LSTM at time $t$ and $Z(h_t)$ is a normalization constant. The LSTM prediction stack acts as a generative process for anticipating future features.

The \textbf{objective function} for training the predictive stack is a weighted zero order hold between the predicted features and the actual observed features, weighted by the zero order hold difference. The prediction error at time $t$ is given by
\begin{equation}
    E(t) = \frac{1}{n_f}\sum^{w_f}_{i=1}\sum^{h_f}_{j=1} \hat{m}_t(i,j)\odot {\lVert f^S_{t+1}(i,j) - \hat{f}^S_{t+1}(i,j)\lVert_{\ell_1}^2}
    \label{eqn:error_func}
\end{equation}
where each feature $f^S_t$ has dimensions $w_f \times h_f \times d_f$ and $\hat{m}_t(i,j)$ is a function that returns the zero order difference between the observed features at times $t$ and $t+1$ at location $(i,j)$. Note that the prediction is done at the feature level and not at the pixel level, which would result in errors at a different granularity. 

\subsection{Prediction Error-based Attention Map}
At the core of our approach is the idea of spatial-temporal prediction error for localizing the actions of interest in the video. It takes into account the quality of the predictions made and the relative alignment of the prediction errors in and around each spatial location. 
The input to the error detection module is the unaveraged result of the objective function from Equation~\ref{eqn:error_func}, which by itself represents the combined spatial-temporal loss. 
We compute a weight $\alpha_{ij}$ associated with each spatial location $(i,j)$ in the predicted feature $\hat{f}^S_{t+1}$ as
\begin{equation}
    \alpha_{ij} = \frac{exp(e_{ij})}{\sum_{m=1}^{w_k}\sum_{n=1}^{h_k}exp(e_{mn})}
    \label{eqn:st_error}
\end{equation}
where $e_{ij}$ represents the weighted prediction error at $(i,j)$ as defined in Equation~\ref{eqn:error_func} and can be considered to be a function $a$ of the state of the top-most LSTM and the input feature $f^S_t$ at time $t$ and can be defined as $a(f^S_t, h^l_{t-1})$. The resulting matrix is an error-based attention map that allows us to localize the prediction error at a specific spatial location where as Equation~\ref{eqn:error_func} allows for temporal localization. 
It is to be noted that this is very similar in formulation of Bahdanau attention~\cite{bahdanau2014neural}. However, there are two key differences. First, our formulation is not parametrized and does not add to the number of learnable parameters in the framework. Second, our attention map is a characterization of the difficulty in anticipating unpredictable motion whereas Bahdanau attention is an effort to increase the encoding ability of the decoder and does not characterize the unpredictability of the future feature. We compare the use of both types of attention in Section~\ref{sec:ablative}, where we see that error-based localization is more suitable for our application.

\subsection{Extraction of Action Tubes}
The action localization module receives a stream of bounding box proposals and an error-based attention map to select an output tube, parametrized as a selection of bounding boxes $\mathcal{T} = \mathcal{B}_{it}$ at each time instance $t$. 
The action localization is a selection algorithm that filters \textit{all} region proposals from Section~\ref{sec:proposal} and returns the collection of proposals that have a higher probability of action localization. We do so by assigning an energy term to each of the bounding box proposals ($\mathcal{B}_{it}$) at time $t$ and choosing the top $k$ bounding boxes with least energy as our final proposals. The energy of a bounding box $\mathcal{B}_{it}$ is defined as 
\begin{equation}
E(\mathcal{B}_{it}) = w_\alpha\ \phi(\alpha_{ij}, \mathcal{B}_{it}) + w_t \delta(\mathcal{B}_{it}, \{\mathcal{B}_{j,t-1}\}) 
\label{eqn:BB_Energy}
\end{equation}
where $\phi(\cdot)$ is a function that returns a value characteristic of the distance between the bounding box center and location of maximum error, $\delta(\cdot)$ is a function that returns the minimum spatial distance between the current bounding box and the closest bounding box from the previous time step and $w_\alpha$ and $w_t$ are scaling factors. Note that $\delta(\cdot)$ is introduced to enforce temporal consistency in predictions, but we find that it is optional since the LSTM prediction stack implicitly enforces the temporal consistency through its memory states. In our experiments we set $k=10, w_\alpha = 0.75$.

\subsection{Implementation Details}
In our experiments, we use a VGG-16~\cite{simonyan2014very} network pre-trained on ImageNet as our feature extraction network. We use the output of the last convolutional layer before the fully connected layers as our spatial features. Hence the dimensions of the spatial features are $w_f=14, h_f=14, d_f=512$. These output features were then used by an SSD~\cite{liu2016ssd} to generate bounding box proposals. Note that we just take the generated bounding box proposals without taking into account classes and associated probabilities. 
We use a three layer hierarchical LSTM model with the hidden state size as $512$ as our predictor module. We use the vanilla LSTM as proposed in ~\cite{hochreiter1997long}. 
Video level-features are obtained by max-pooling the element-wise dot-product of hidden state of the top-most LSTM and the attention values across time.
We train with the adaptive learning mechanism proposed in ~\cite{Aakur_2019_CVPR}, with the initial learning rate set to be $1\times10^{-8}$ and scaling factors $\Delta^{-}_{t}$ and $\Delta^{+}_{t}$ as $1\times10^{-2}$ and $1\times10^{-3}$, respectively. The network was trained for $1$ epoch on a computer with one Titan X Pascal. 

\section{Experimental Setup}
\subsection{Data}
We evaluate our approach on three publicly available datasets for evaluating the proposed approach on the action localization task. 

\textbf{UCF Sports}~\cite{rodriguez2008action} is an action localization dataset consisting of $10$  classes of sports actions such as skating and lifting collected from sports broadcasts. It is an interesting dataset since it has a high concentration of distinct scenes and motions that make it challenging for localization and recognition. We use the splits ($103$ training and $47$ testing videos) as defined in ~\cite{lan2011discriminative} for evaluation.

\textbf{JHMDB}~\cite{jhuang2013towards} is composed of $21$ action classes and $928$ trimmed videos. All videos are annotated with human-joints for every frame. The ground truth bounding box for the action localization task is chosen such that the box encompasses all the joints.  This dataset offers several challenges, such as increasing amounts of background clutter, high inter-class similarity, complex motion (including camera motion), and occluded objects of interest. We report all results as the average across all three splits.

\textbf{THUMOS'13}~\cite{jiang2014thumos} is a subset of the UCF-101~\cite{soomro2012ucf101} dataset, consisting of $24$ classes and $3,207$ videos. Ground truth bounding boxes are provided for each of the classes for the action localization task. It is also known as the \textbf{UCF-101-24} dataset. Following prior works~\cite{soomro2017unsupervised,li2018videolstm}, we perform our experiments and report results on the first split.

We also evaluate the generalization ability of the proposed approach on egocentric videos by evaluating on the \textit{unsupervised gaze prediction task}. There has been evidence from cognitive psychology that there is a strong correlation between gaze points and action localization~\cite{tipper1992selective}. Hence, the gaze prediction task would be a reasonable measure of the generalization to action localization in egocentric videos. We evaluate the performance on the \textbf{GTEA Gaze}~\cite{fathi2012learning} dataset, which consists of $17$ sequences of tasks performed by $14$ subjects, with each sequence lasting about $4$ minutes. We use the official splits for the GTEA datasets as defined in prior works ~\cite{fathi2012learning}. 

\subsection{Metrics and Baselines}\label{sec:metric}
For the \textbf{action localization} task, we follow prior works~\cite{soomro2017unsupervised,li2018videolstm} and report the mean average precision (mAP) at various overlap thresholds, obtained by computing the Intersection Over Union (IoU) of the predicted and ground truth bounding boxes. We also evaluate the quality of bounding box proposals by measuring the average, per-frame IoU, and the bounding box \textit{recall} at varying overlap ratios. 

Since ours is an unsupervised approach, we obtain class labels by clustering the learned representations using the \textit{k-means} algorithm. While more complicated clustering may yield better recognition results~\cite{soomro2017unsupervised}, the k-means approach allows us to evaluate the robustness of learned features. We evaluate our approach in two settings $K_{gt}$ and $K_{opt}$, where the number of clusters is set to the number of ground truth action classes and an optimal number obtained through the elbow method~\cite{kodinariya2013review}, respectively. From our experiments, we observe that $K_{opt}$ is typically three times the number of ground truth classes, which is not unreasonable and has been a working assumption in other deep learning-based clustering approaches~\cite{hershey2016deep}.  

We also compare against other LSTM and attention-based approaches (Section~\ref{sec:lstm}) to the action localization problem for evaluating the effectiveness of the proposed training protocol.

For the \textbf{gaze prediction} task, we evaluate the approaches using \textbf{Area Under the Curve} (AUC), which measures the area under the curve on saliency maps for true positive versus false-positive rates under various threshold values. We also report the \textbf{Average Angular Error} (AAE), which measures the angular distance between the predicted and ground truth gaze positions. 
Since the output of our model is a saliency map, AUC is a more appropriate metric compared to average angular error (AAE), which requires specific locations.

\section{Quantitative Evaluation}
In this section, we present the quantitative evaluation of our approach on two different tasks, namely action localization, and egocentric gaze prediction. For the action localization task, we evaluate our approach on two aspects - the quality of proposals and spatial-temporal localization.

\subsection{Quality of Localization Proposals}
We first evaluate the quality of our localization proposals by assuming perfect class prediction. This allows us to independently assess the quality of localization performed in a self-supervised manner. We present the results of the evaluation in Table~\ref{table:perf} and compare against fully supervised, weakly supervised, and unsupervised baselines. As can be seen, we outperform many supervised and weakly supervised baselines. APT~\cite{van2015apt} achieves a higher localization score. However, it produces, on average, $1,500$ proposals per video, whereas our approach returns approximately $10$ proposals. A large number of localization proposals per video can lead to higher recall and IoU but makes the localization task i.e., action labeling per video harder and can affect the ability to generalize across domains. 
Also, it should be noted that our approach produces proposals in \textit{streaming} fashion, as opposed to many of the other approaches, which produce action tubes based on motion computed across the entire video. This can make real-time action localization in streaming videos harder. 
\begin{table}
\centering
\begin{tabular}{|c|c|c|}
\hline
\multirow{2}{*}{\textbf{Supervision}} & \multirow{2}{*}{\textbf{Approach}}      & \multirow{2}{*}{\textbf{Average}}\\ 
  &    & \\ 
  \toprule
    \multirow{2}{*}{Full} 
& STPD\cite{tran2011optimal} & 44.6 \\
& Max Path Search~\cite{tran2012max} & \textbf{54.3} \\
 \midrule
     \multirow{3}{*}{Weak} 
& Ma et al~\cite{ma2013action} & 44.6 \\
& GBVS~\cite{grundmann2010efficient} & 42.1 \\
& Soomro et al~\cite{soomro2017unsupervised} & \textbf{47.7} \\
 \midrule
 \multirow{2}{*}{None} 
& IME Tublets~\cite{jain2014action} & 51.5 \\
& APT~\cite{van2015apt} & \textbf{63.7}\\
& Proposed Approach & 55.7 
\\\bottomrule
\end{tabular}
\caption{Comparison with fully supervised and weakly supervised baselines on class-agnostic action localization on UCF Sports dataset.}
\label{table:perf}
\end{table}

\subsection{Spatial-temporal Action Localization}
\begin{figure*}
\begin{tabular}{ccc}
  \includegraphics[width=0.32\textwidth]{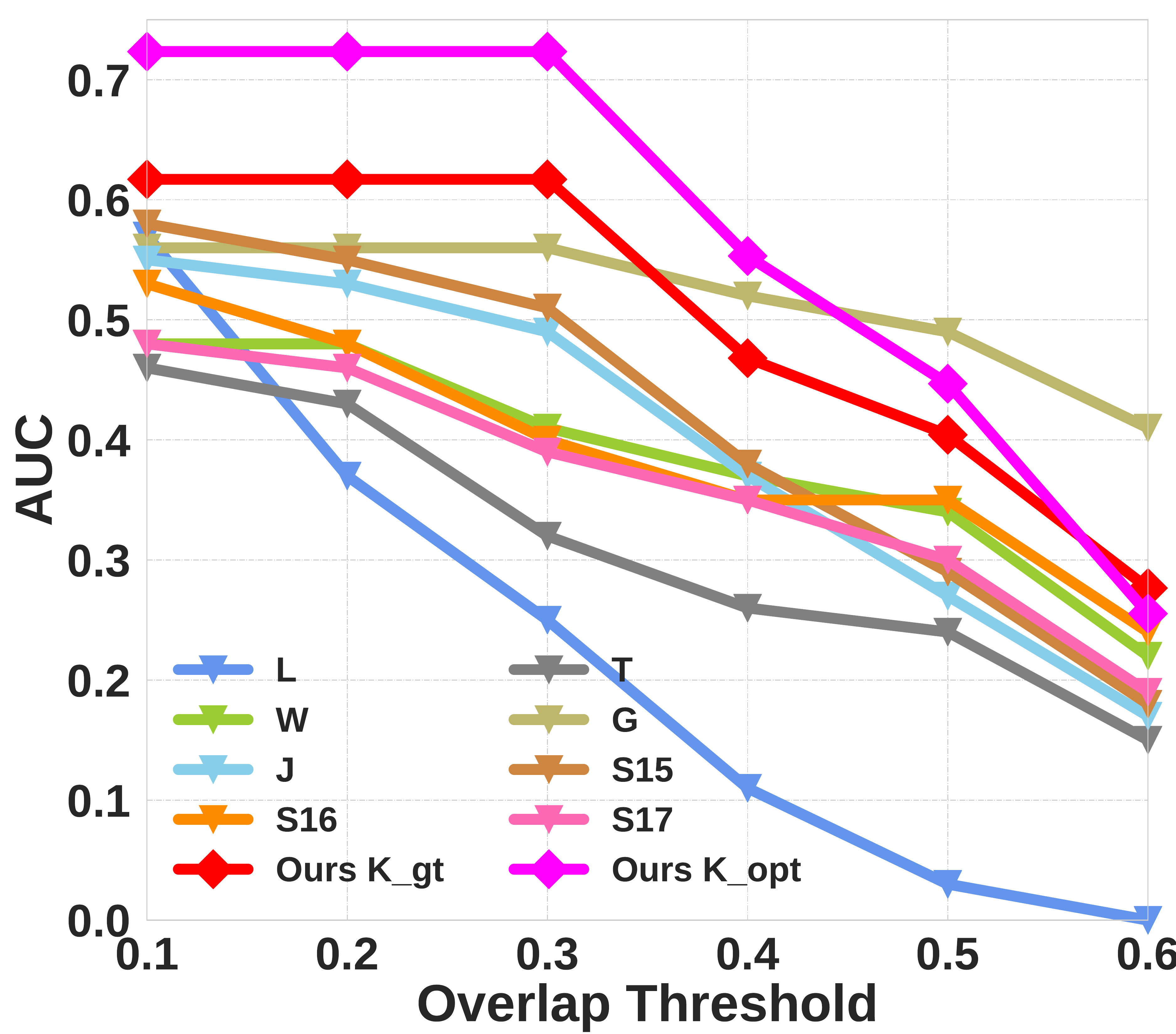} & \includegraphics[width=0.32\textwidth]{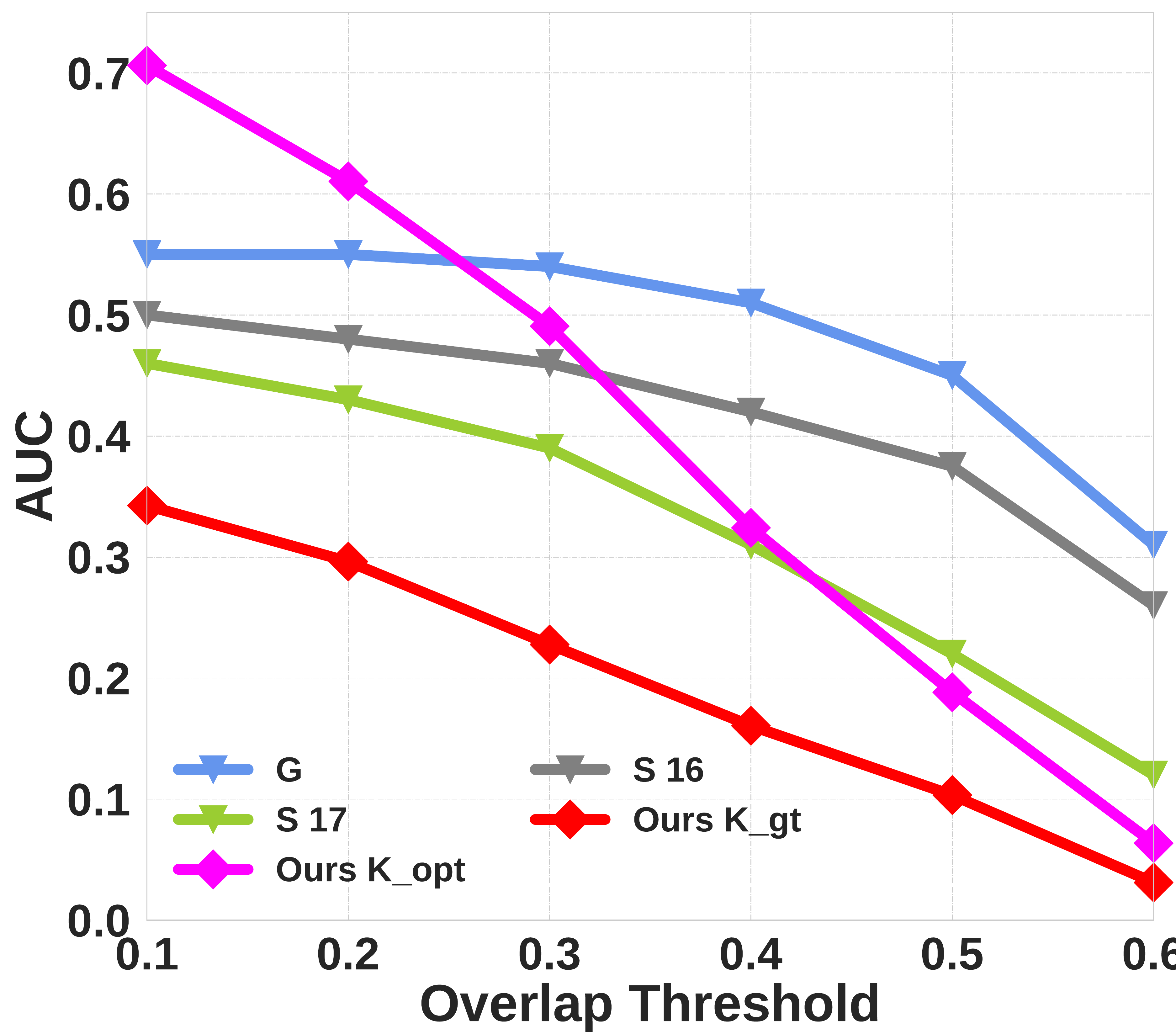} & 
  \includegraphics[width=0.32\textwidth]{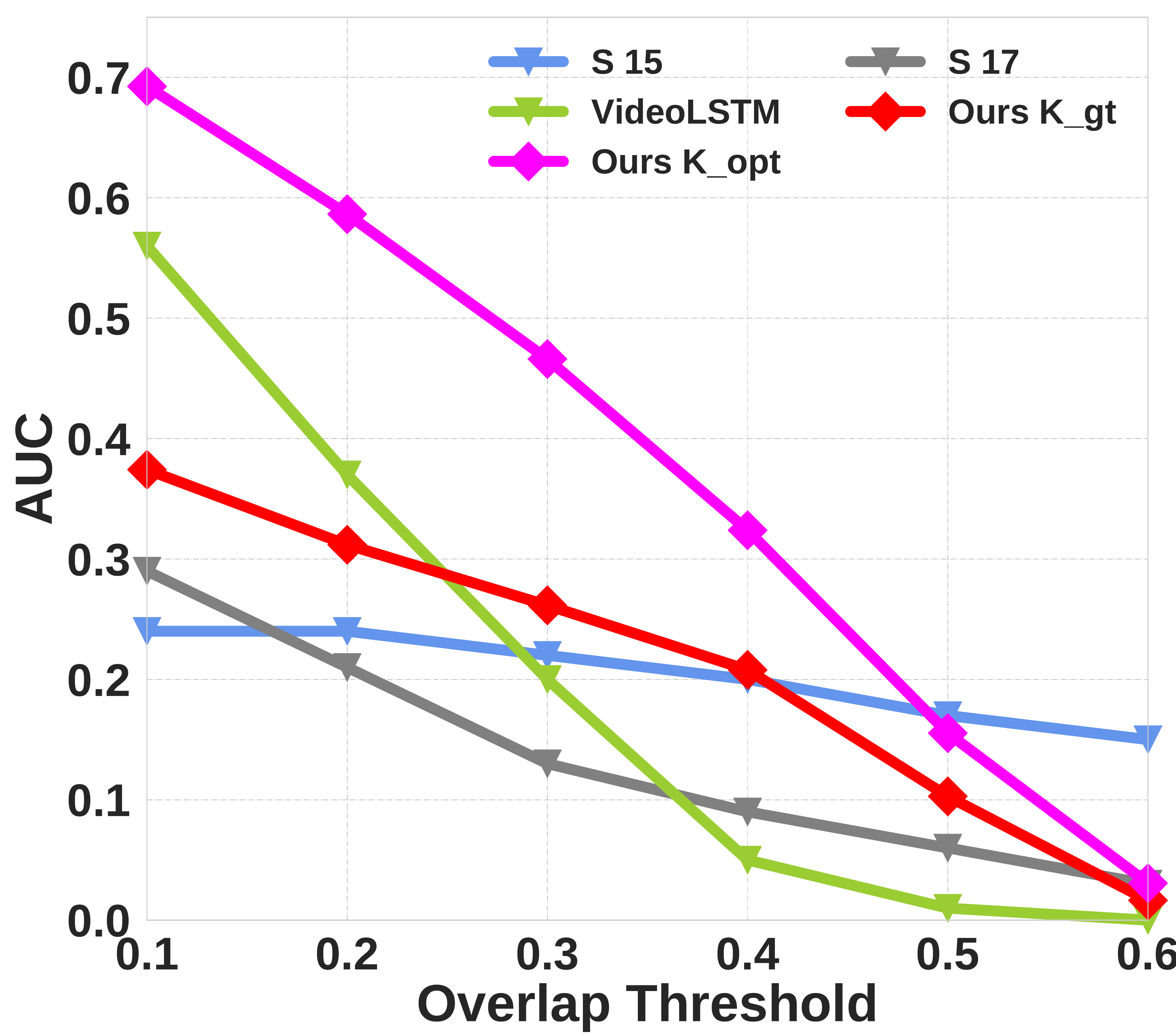} \\
            (a) & (b) & (c)\\
\end{tabular}
\caption{AUC for the action localization tasks are shown for (a) UCF Sports, (b) JHMDB and (c) THUMOS13 datasets. We compare against baselines with varying levels of supervision such as \textbf{L}an \textit{et al.}~\cite{lan2011discriminative}, \textbf{T}ian \textit{et al.}~\cite{tian2013spatiotemporal}, \textbf{W}ang \textit{et al.}~\cite{wang2014video}, \textbf{G}kioxari and Malik~\cite{gkioxari2015finding}, \textbf{J}ain \textit{et al.}~\cite{jain2014action}, \textbf{S}oomro \textit{et al}~\cite{soomro2016predicting,soomro2015action,soomro2017unsupervised} and VideoLSTM~\cite{li2018videolstm}.}
\label{fig:auc}
\end{figure*}

We also evaluate our approach on the spatial-temporal localization task. This evaluation allows us to analyze the robustness of the self-supervised features learned through prediction. 
We generate video-level class labels through clustering and use the standard evaluation metrics (Section ~\ref{sec:metric}) to quantify the performance. 
The AUC curves with respect to varying overlap thresholds are presented in Figure~\ref{fig:auc}. We compare against a mix of supervised, weakly-supervised, and unsupervised baselines on all three datasets. 

On the \textbf{UCF Sports} dataset (Figure~\ref{fig:auc}(a)), we outperform all baselines including several supervised baselines except for Gkioxari and Malik~\cite{gkioxari2015finding} at higher overlap thresholds ($\sigma > 0.4$) when we set number of clusters $k$ to the number of ground truth classes. When we allow for some over-segmentation and use the \textit{optimal} number of clusters, we outperform all baselines till $\sigma > 0.5$.

On the \textbf{JHMDB} dataset (Figure~\ref{fig:auc}(b)), we find that our approach, while having high recall ($77.8\% @ \sigma=0.5$), the large camera motion and intra-class variations have a significant impact on the classification accuracy. Hence, the mAP suffers when we set $k$ to be the number of ground truth classes. When we set the number of clusters to the optimal number of clusters, we outperform other baselines at lower thresholds ($\sigma<0.5$). It should be noted that the other unsupervised baseline (Soomro \textit{et al}~\cite{soomro2017unsupervised}) uses object detection proposals from a Faster R-CNN backbone to score the "\emph{humanness}" of a proposal. This assumption tends to make the approach biased towards human-centered action localization and affects its ability to generalize towards actions with non-human actors. We, on the other hand, do not make any assumptions on the characteristics of the actor, scene, or any motion dynamics. 

On the \textbf{THUMOS'13} dataset (Figure~\ref{fig:auc}(c)), we achieve consistent improvements over unsupervised and weakly supervised baselines, at $k=k_{gt}$ and achieve state-of-the-art mAP scores when $k=k_{opt}$. It is interesting to note that we perform competitively (when $k=k_{gt}$) the weakly-supervised attention-based VideoLSTM~\cite{li2018videolstm}, which uses a convLSTM for temporal modeling along with a CNN-based spatial attention mechanism. It should be noted that we have a higher recall rate ($0.47@\sigma=0.4$ and $0.33@\sigma=0.5$)  at higher thresholds compared to other state-of-the-art approaches on THUMOS'13 and shows the robustness of the error-based localization approach to intra-class variation and occlusion. 

\textbf{Clustering quality.} Since there is a significant difference in the mAP score when we set a different number of clusters in k-means, we measured the homogeneity (or purity) of the clustering. The homogeneity score measures the ``quality'' of the cluster by measuring how well a cluster models a given ground-truth class. Since we allow the over-segmentation of clusters when we set $k$ to the optimal number of clusters, this is an important measure of feature robustness. Higher homogeneity indicates that intra-class variations are captured since all data points in a given cluster belong to the same ground truth class. We observe an average homogeneity score of 74.56\% when $k$ is set to the number of ground truth classes and 78.97\% when we use the optimal number of clusters. As can be seen, although we over-segment, each of the clusters typically models a single action class to a high degree of integrity. 

\subsection{Comparison with other LSTM-based approaches}\label{sec:lstm}
We also compare our approach with other LSTM-based and attention-based models to highlight the importance of the proposed self-supervised learning paradigm. 
Since LSTM-based frameworks can have highly similar architectures, we take into account different requirements and characteristics, such as the level of annotations required for training and the number of localization proposals returned per video. 
We compare with three approaches similar in spirit to our approach - ALSTM~\cite{sharma2015action}, VideoLSTM~\cite{li2018videolstm} and  Actor Supervision~\cite{escorcia2020guess} and summarize the results in Table~\ref{table:compare_lstm}. It can be seen that we significantly outperform VideoLSTM and ALSTM on the THUMOS'13 dataset in both recall and $mAP@\sigma=0.2$. 
\begin{table}
\centering
\begin{tabular}{|c|c|c|c|c|c|c|c|c|c|c|}
\hline

\multirow{2}{*}{Approach} & \multicolumn{2}{|c|}{Annotations} & \multirow{2}{*}{\# Proposals} & \multicolumn{5}{|c|}{Average Recall}  & mAP\\
\cline{2-3}
\cline{5-9}
  & Labels & Boxes & & $0.1$ & $0.2$ & $0.3$ & $0.4$ & $0.5$ & $@0.2$\\ 
  \toprule
ALSTM~\cite{sharma2015action} &  \ding{51} & \ding{55} & 1 & 0.46 & 0.28 & 0.05 & 0.02 & - & 0.06 \\\hline
VideoLSTM~\cite{li2018videolstm} &  \ding{51}  & \ding{55} & 1 & 0.71 & 0.52 & 0.32 & 0.11 & - & 0.37\\ \hline
Actor Supervision*~\cite{escorcia2020guess}& \ding{51} & \ding{55} & $\sim1000$ & \textbf{0.89} & - & - & -  & \textbf{0.44} & 0.46 \\\hline
Proposed Approach& \ding{55} & \ding{55} & $\sim10$ & 0.84 & \textbf{0.72} & \textbf{0.58} & \textbf{0.47} & 0.33 & \textbf{0.59}\\
\bottomrule
\end{tabular}
\caption{Comparison with other LSTM-based and attention-based approaches on the THUMOS'13 dataset. We report average recall at various overlap thresholds, mAP at $0.2$ overlap threshold and the average number of proposals per frame. 
}
\label{table:compare_lstm}
\end{table}
Actor Supervision~\cite{escorcia2020guess} outperforms our approach on recall, but it is to be noted that the region proposals are dependent on two factors - (i) object detection-based actor proposals and (ii) a filtering mechanism that limits proposals based on ground truth action classes, which can increase the training requirements and limit generalizability. Also, note that returning a higher number of localization proposals can increase recall at the cost of generalization. 

\subsection{Ablative Studies}\label{sec:ablative}
The proposed approach has three major units that affect its performance the most - (i) the region proposal module, (ii) future prediction module, and (iii) error-based action localization module. In our primary framework, we use the prediction error-based localization to choose bounding box proposals generated through a class-agnostic SSD object detection model. We make the SSD class-agnostic by only considering bounding box proposals at all detection thresholds. 
We consider and evaluate several alternatives to all three modules. We choose selective search~\cite{uijlings2013selective} and EdgeBox~\cite{zhu2015tracking} as alternative region proposal methods. 
We use an attention-based localization method for action localization as an approximation of the ALSTM~\cite{sharma2015action} to evaluate the effectiveness of using the proposed error-based localization. 

We also evaluate the effect of attention-based prediction by introducing a Bahdanau~\cite{bahdanau2014neural} attention layer before prediction. 
We evaluate these approaches on the UCF Sports dataset and visualize the results in Figure~\ref{fig:class_auc}(c). It can be seen that the use of the prediction error-based localization has a significant improvement over a trained attention-based localization approach. We can also see that the choice of region proposal methods do have some effect on the performance of the approach, with selective search and EdgeBox proposals doing slightly better at higher thresholds ($\sigma \in (0.4, 0.5)$) at the cost of inference time and additional bounding box proposals ($50$ compared to the $10$ from SSD-based region proposal). Using SSD for generating proposals allows us to share weights across the frame encoder and region proposal tasks and hence reduce the memory and computational footprint of the approach. 
We also find that using attention as part of the prediction module significantly impacts the performance of the architecture. It could, arguably, be attributed to the objective function, which aims to minimize the prediction error. Using attention to encode the input could impact the prediction function. 

\subsection{Unsupervised Egocentric Gaze Prediction}
Finally, we evaluate the ability to generalize to egocentric videos by quantifying the performance of the model on the unsupervised gaze prediction task. Given that we do not need any annotations or other auxiliary data, we employ the same architecture and training strategy for this task. We evaluate on the GTEA gaze dataset and compare it with other unsupervised models in Table~\ref{table:egocentric}. 
\begin{table}
\centering
\begin{tabular}{|c|c|c|c|c|c|c|}
\hline
 & Itti \textit{et al.}~\cite{itti2000saliency} & GBVS~\cite{harel2007graph} & AWS-D~\cite{leboran2016dynamic} & Center Bias & OBDL~\cite{hossein2015many} & Ours \\
 \toprule
AUC & 0.747 & 0.769 & 0.770 & 0.789 & 0.801  & \textbf{0.861} \\\midrule
AAE & 18.4 & 15.3  & 18.2  & \textbf{10.2} & 15.6  & 13.6 \\
\bottomrule
\end{tabular}
\caption{Comparison with state-of-the-art on the unsupervised egocentric gaze prediction task on the GTEA dataset.}
\label{table:egocentric}
\end{table}
As can be seen, we obtain competitive results on the gaze prediction task, outperforming all baselines on both the AUC and AAE scores. It is to be noted that we outperform the center bias method on the AUC metric. Center bias exploits the spatial bias in egocentric images and always predicts the center of the video frame as the predicted gaze position. The significant improvement in the AUC metric indicates that our approach predicts gaze fixations that are more closely aligned with the ground truth than the center bias approach. Although we do not return a specific gaze position, we outperform all baselines except center bias on the AAE metric. Given that the model was not designed explicitly for this task, it is a remarkable performance, especially given the performance of fully supervised baselines such as DFG~\cite{zhang2017deep}, which achieve $10.6$ and $88.3$ for AUC and AAE.

\subsection{Qualitative Evaluation}
\begin{figure*}
\centering
\begin{tabular}{cccccc}
\toprule
\multicolumn{6}{c}{Successful Localization}\\
\toprule
 \includegraphics[width=0.14\textwidth]{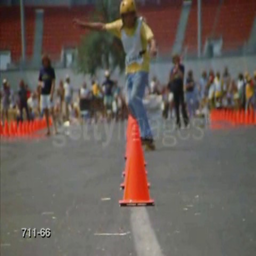} &
  \includegraphics[width=0.14\textwidth]{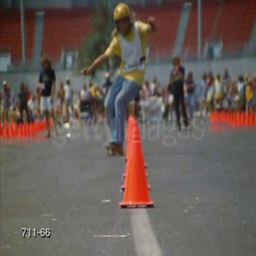} & \includegraphics[width=0.14\textwidth]{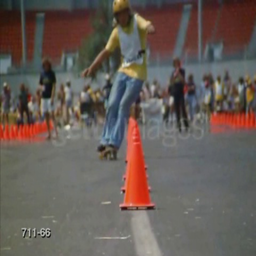} & 
  \includegraphics[width=0.14\textwidth]{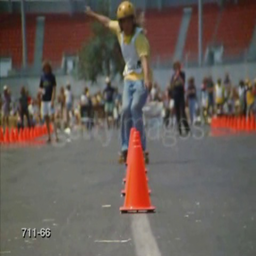} & \includegraphics[width=0.14\textwidth]{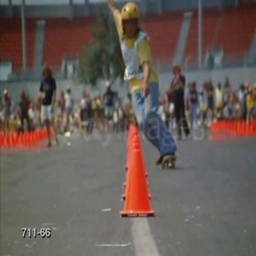} & 
  \includegraphics[width=0.14\textwidth]{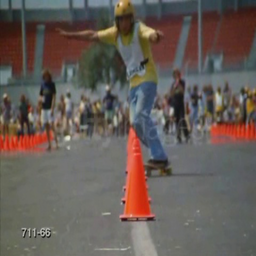} \\

\includegraphics[width=0.14\textwidth]{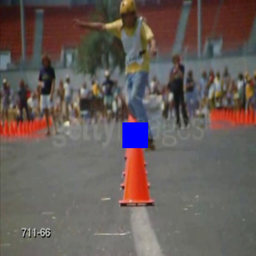} &
  \includegraphics[width=0.14\textwidth]{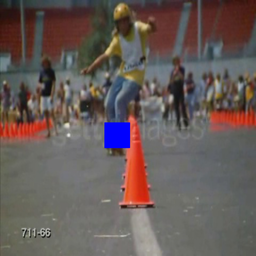} & \includegraphics[width=0.14\textwidth]{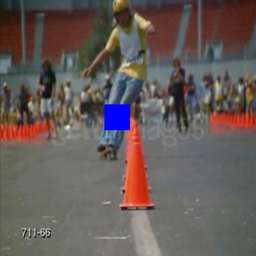} & 
  \includegraphics[width=0.14\textwidth]{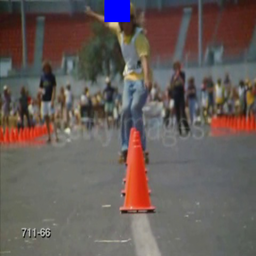} & \includegraphics[width=0.14\textwidth]{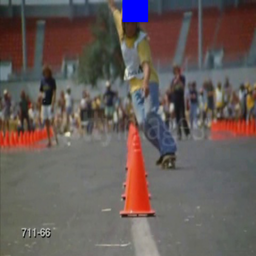} & 
  \includegraphics[width=0.14\textwidth]{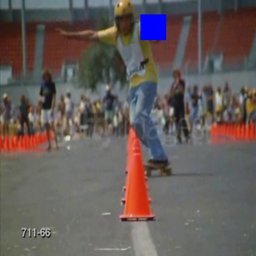} \\

\includegraphics[width=0.14\textwidth]{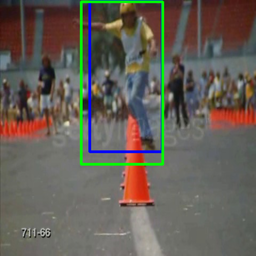} &
  \includegraphics[width=0.14\textwidth]{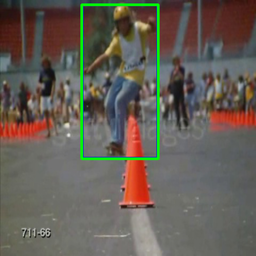} & \includegraphics[width=0.14\textwidth]{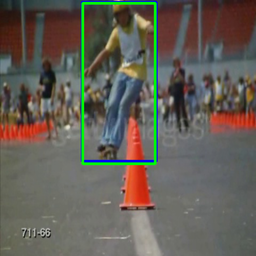} & 
  \includegraphics[width=0.14\textwidth]{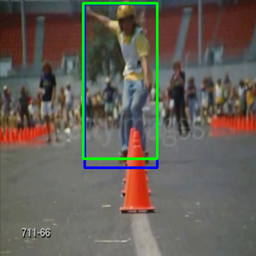} & \includegraphics[width=0.14\textwidth]{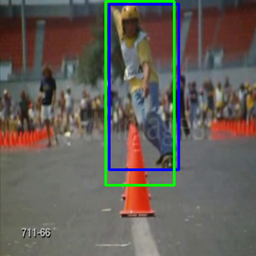} & 
  \includegraphics[width=0.14\textwidth]{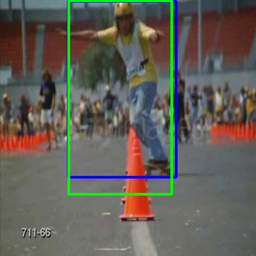} \\
 
\toprule
\multicolumn{6}{c}{Unsuccessful Localization}\\
\toprule
\includegraphics[width=0.15\textwidth]{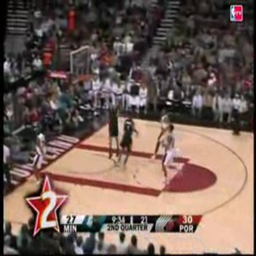} &
  \includegraphics[width=0.15\textwidth]{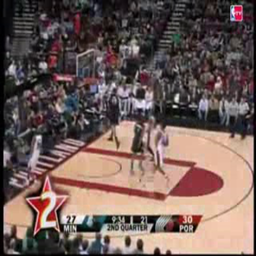} & 
  \includegraphics[width=0.15\textwidth]{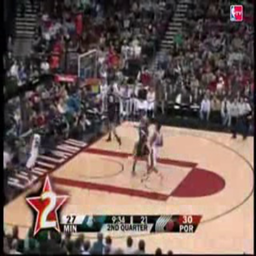} & 
  \includegraphics[width=0.15\textwidth]{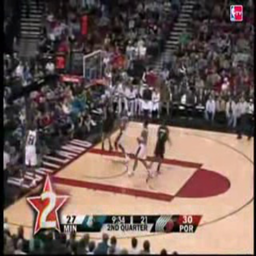} & 
  \includegraphics[width=0.15\textwidth]{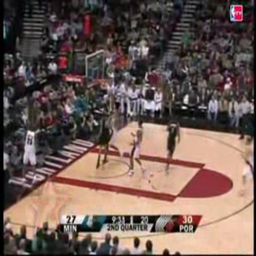} & 
  \includegraphics[width=0.15\textwidth]{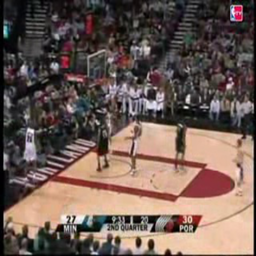} \\

\includegraphics[width=0.15\textwidth]{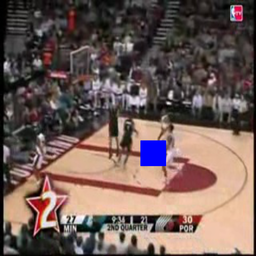} &
  \includegraphics[width=0.15\textwidth]{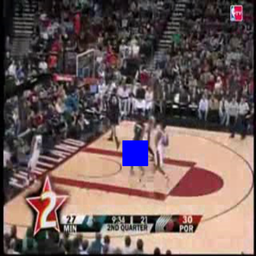} & 
  \includegraphics[width=0.15\textwidth]{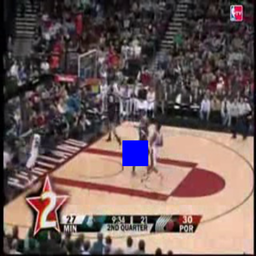} & 
  \includegraphics[width=0.15\textwidth]{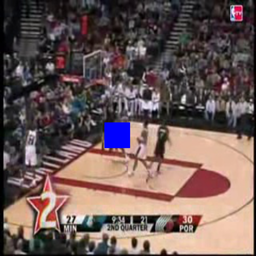} & 
  \includegraphics[width=0.15\textwidth]{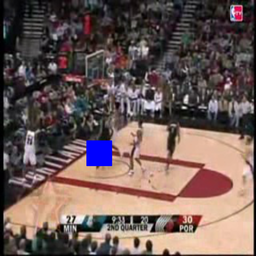} & 
  \includegraphics[width=0.15\textwidth]{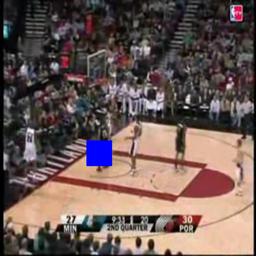} \\

\includegraphics[width=0.15\textwidth]{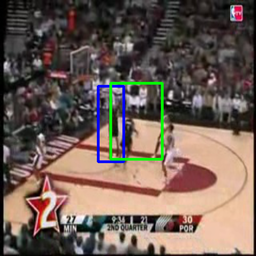} &
  \includegraphics[width=0.15\textwidth]{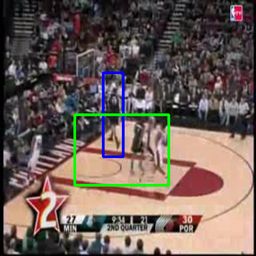} & 
  \includegraphics[width=0.15\textwidth]{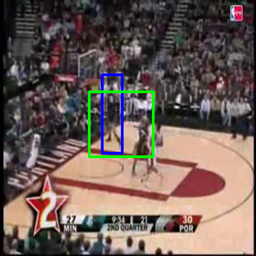} & 
  \includegraphics[width=0.15\textwidth]{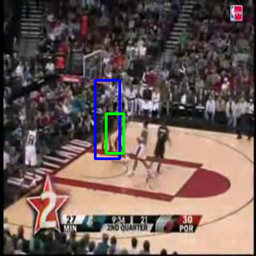} & 
  \includegraphics[width=0.15\textwidth]{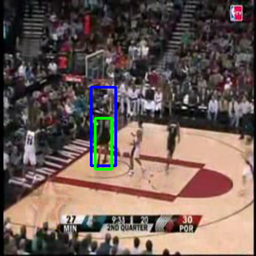} & 
  \includegraphics[width=0.15\textwidth]{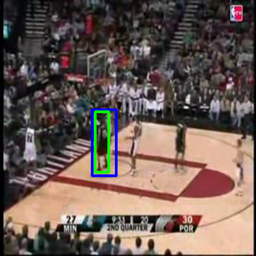} \\
 
\end{tabular}
\caption{\textbf{Qualitative Examples}: We present the input frame, error-based attention location and the final prediction,  
for both successful and unsuccessful localizations. 
Green BB: Prediction, Blue BB: Ground truth}
\label{fig:qual_res}
\end{figure*}

We also qualitatively analyze the proposed approach to identify its strengths and weaknesses. We find that our approach has a consistently high recall for the localization task across datasets and domains. We consider that an action is correctly localized if the average IoU across all frames is higher than $0.5$, which indicates that most, if not all, frames in a video are correctly localized. 
\begin{figure*}
\begin{tabular}{ccc}
  \includegraphics[width=0.32\textwidth]{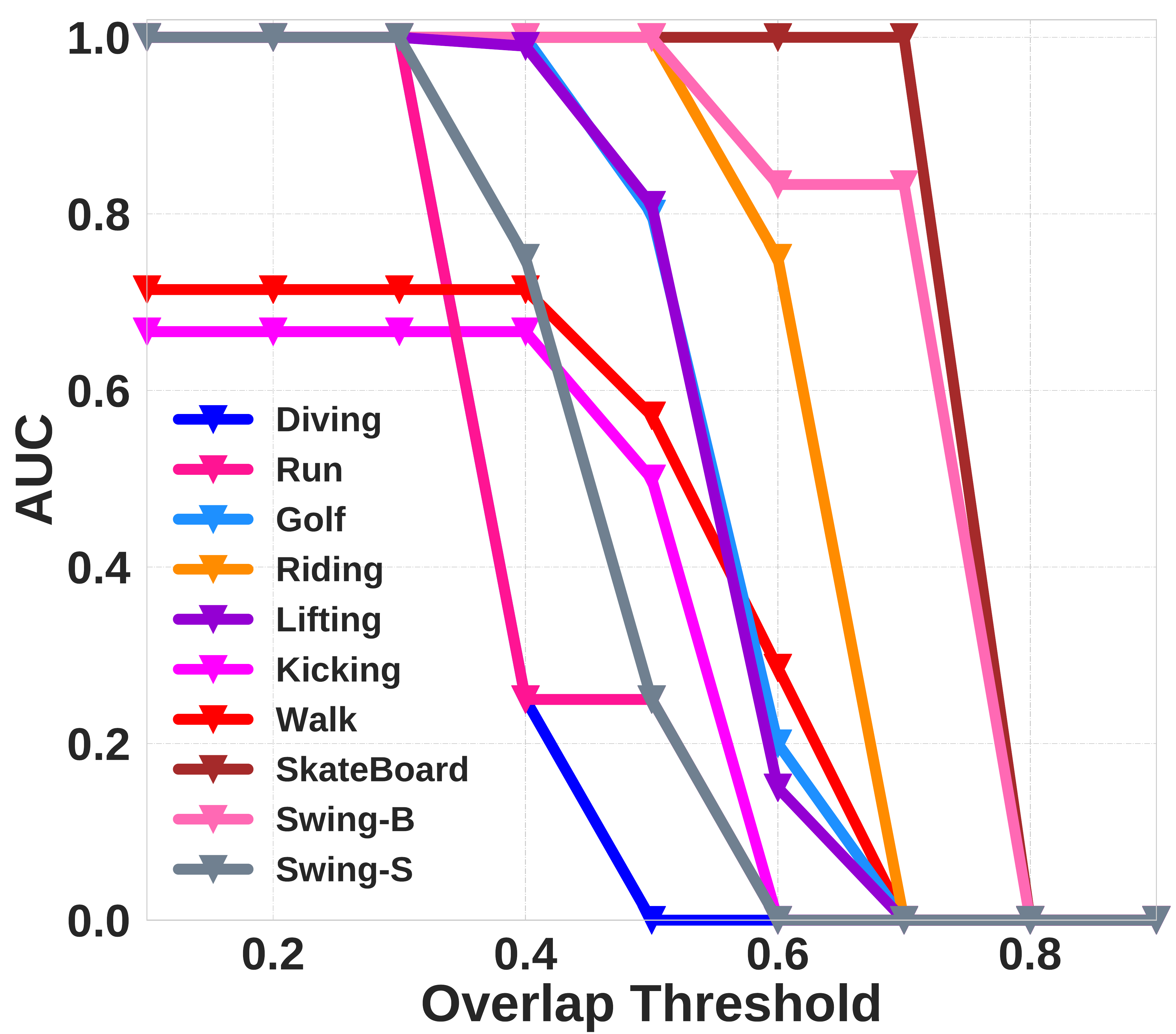} & \includegraphics[width=0.32\textwidth]{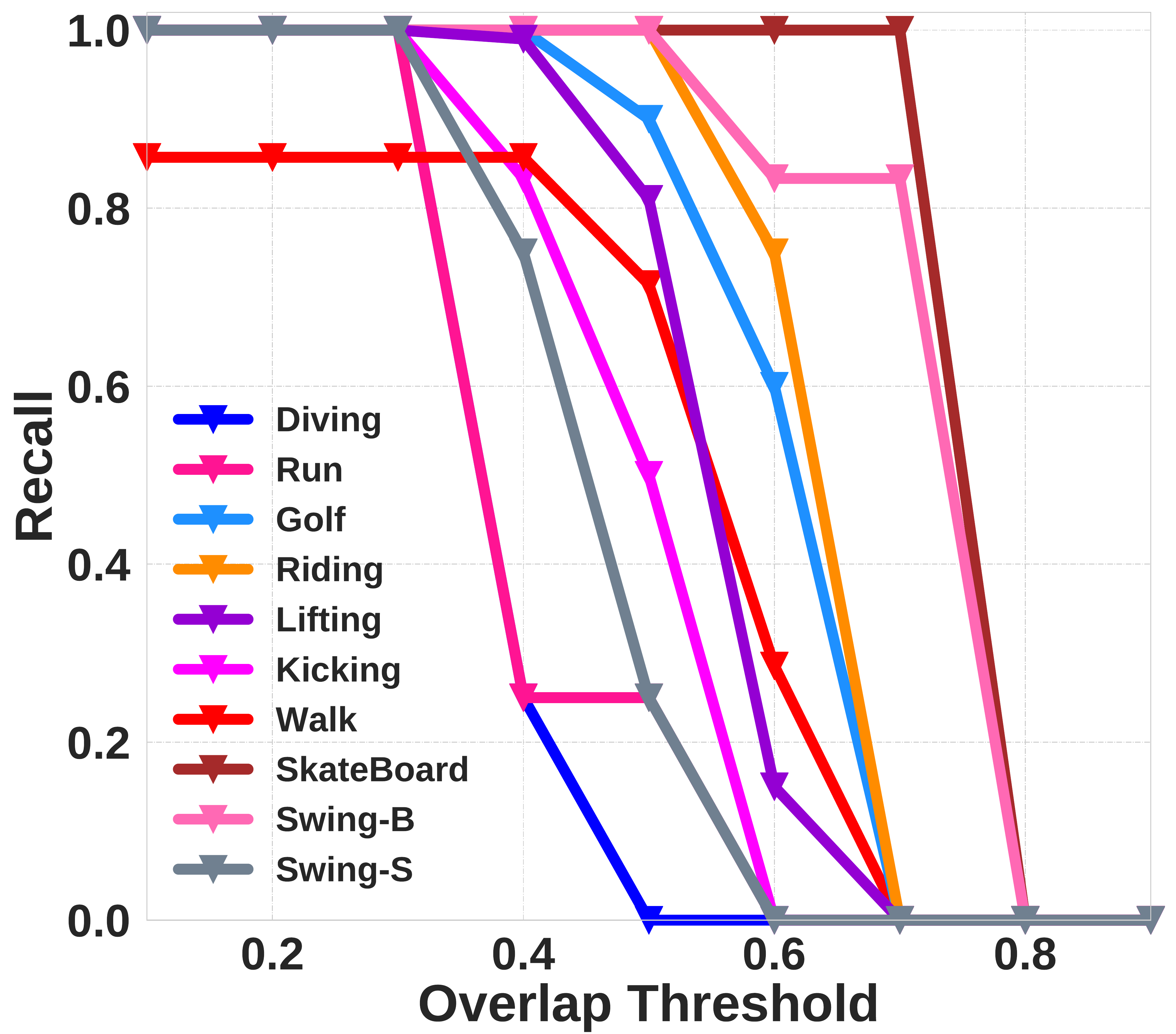} & 
  \includegraphics[width=0.32\textwidth]{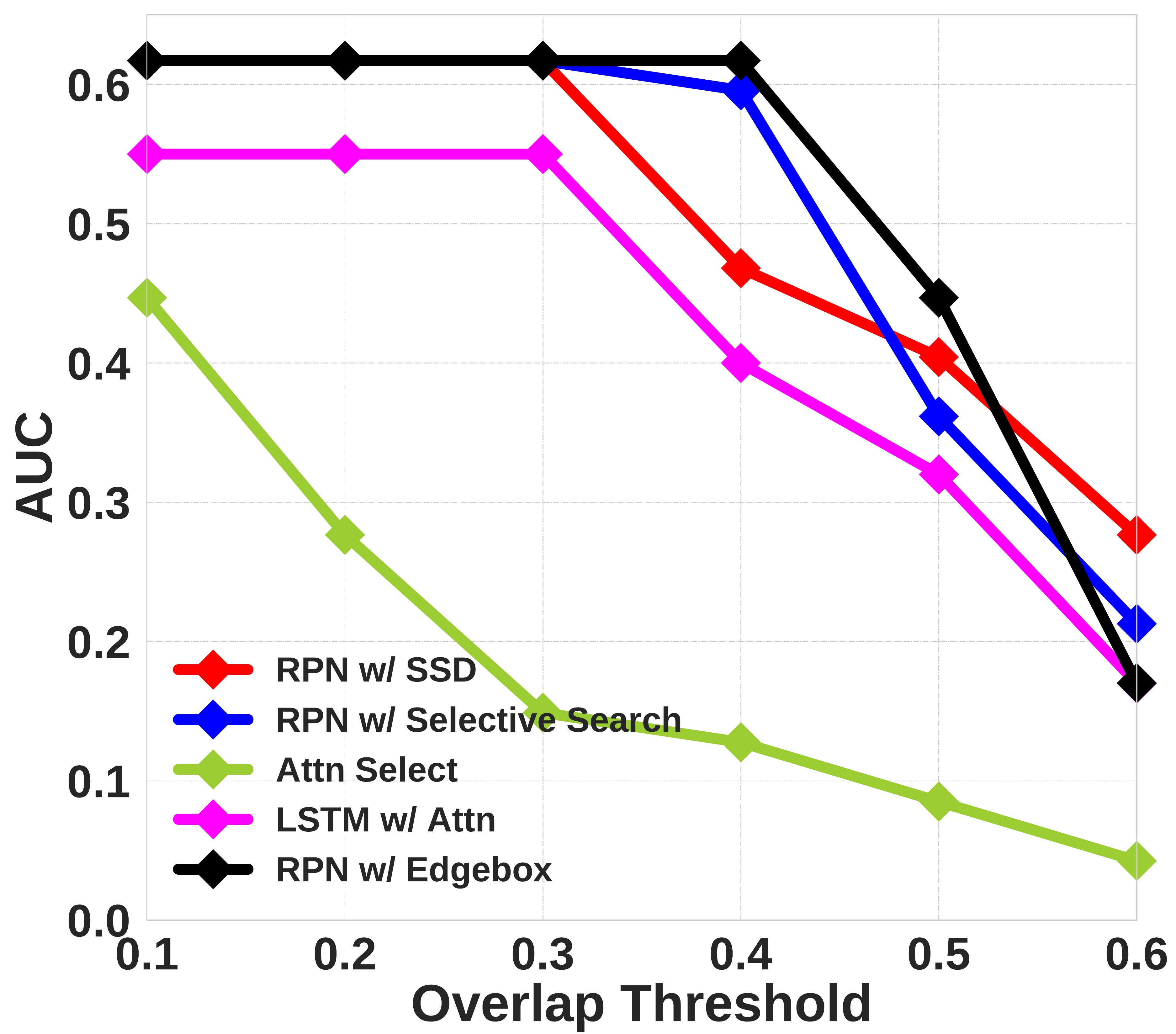} \\
            (a) & (b) & (c)\\
            
\end{tabular}
\caption{Qualitative analysis of the proposed approach on UCF Sports dataset by illustrating (a) class-wise AUC, (b) class-wise bounding box recall at different overlap thresholds and (c) effect of each module on AUC.}
\label{fig:class_auc}
\end{figure*}
We illustrate the recall scores and subsequent AUC scores for each class in the UCF sports dataset in Figure~\ref{fig:class_auc}(a-b). It can be seen that for many classes ($7/10$ to be specific), we have more than $80\%$ recall at an overlap threshold of $0.5$. We find, through visual inspection, that the spatial-temporal error is often correlated with the actor, but is often not at the center of the region of interest and thus reduces the quality of the chosen proposals. We illustrate this effect in Figure~\ref{fig:qual_res}. The first row shows the input frame, the second shows the error-based attention, and the last row shows the final localization proposals. If more proposals are returned (as is the case with selective search and EdgeBox), we can obtain a higher recall (Figure~\ref{fig:class_auc}(b)) and higher mAP. More qualitative results in the supplementary.
\section{Conclusion}
In this work, we introduce a self-supervised approach to action localization, driven by spatial-temporal error localization. We show that the use of self-supervised prediction using video frames can help learn highly robust features and obtain state-of-the-art results on localization without any training annotations. We also show that the proposed framework can work with a variety of proposal generation methods 
without losing performance. We also show that the approach can generalize to egocentric videos without changing the training methodology or the framework and obtain competitive performance on the unsupervised gaze prediction task.

\clearpage

\bibliographystyle{splncs04}
\bibliography{egbib}
\end{document}